\documentclass[a4paper,fleqn]{cas-dc}
\usepackage[authoryear]{natbib} 

\usepackage{threeparttable}
\usepackage{xcolor}
\usepackage{graphicx}

\usepackage{subcaption}
\usepackage{tabularx}

\def\tsc#1{\csdef{#1}{\textsc{\lowercase{#1}}\xspace}}
\tsc{WGM}
\tsc{QE}

\newdefinition{rmk}{Remark}
\usepackage{amsmath}
\usepackage{upgreek}
\usepackage[pagewise]{lineno}
\usepackage{float}
\usepackage{placeins}
\usepackage{flushend}
\begin{document}
\begin{sloppypar} 

\let\WriteBookmarks\relax
\def\floatpagepagefraction{1}
\def\textpagefraction{.001}
\let\printorcid\relax

\makeatletter
\renewcommand{\fnum@figure}{Fig. \thefigure.\@gobble}
\makeatother

\shorttitle{}    

\shortauthors{}  

\newcommand{\method}{\textsc{Flavor}}
\newcolumntype{Y}{>{\raggedright\arraybackslash}X}

\title[mode=title]{Beyond Memory Limits: A Hierarchical Spatio-Temporal Map Representation for Large-Scale 3D Autonomous Exploration}

\tnotetext[1]{This research was jointly supported by the National Natural Science Foundation of China (Nos. 42201477 and 42130105) and the Wuhan Natural Science Foundation (No. 2025041001010363).}

\author[1]{Yizhe Zhang}

\ead{yizhezhang0418@whu.edu.cn}
\credit{Conceptualization of this study, Methodology, Software, Data acquisition, Equipment maintenance, Validation, Writing -- original draft}
\affiliation[1]{organization={State Key Laboratory of Information Engineering in Surveying, Mapping and Remote Sensing},
            addressline={Wuhan University}, 
            city={Wuhan},
            postcode={430079}, 
            country={China}}

\author[2]{Jianping Li}
\cormark[1]
\ead{jianping.li@ntu.edu.sg}
\credit{Conceptualization of this study, Methodology, Equipment maintenance, Writing -- review and editing, Funding acquisition}
\affiliation[2]{organization={School of Electrical and Electronic Engineering},
            addressline={Nanyang Technological University}, 
           postcode={639798}, 
            country={Singapore}}

\author[3]{Yilun Gao}
\ead{2024302121458@whu.edu.cn}
\credit{Data acquisition, Equipment maintenance, Validation}
\affiliation[3]{organization={Electronic Information School},
            addressline={Wuhan University},
            city={Wuhan},
            postcode={430072},
            country={China}}

\author[1]{Liangliang Yin}
\ead{yll_whu@whu.edu.cn}
\credit{Data acquisition, Equipment maintenance, Validation}

\author[2]{Pengfei Wan}
\ead{pengfei008@e.ntu.edu.sg}
\credit{Equipment maintenance, Validation}

\author[1]{Bisheng Yang}
\ead{bshyang@whu.edu.cn}
\credit{Conceptualization of this study, Project administration, Funding acquisition}

\cortext[1]{Corresponding author}

\begin{abstract}
Autonomous UAV exploration is increasingly used for large-scale surveying and inspection in inaccessible or disaster-affected environments, where onboard memory and computational resources are often severely constrained. Most existing UAV exploration systems maintain globally queryable geometric map representations for motion planning and exploration decision-making, causing both memory consumption and spatial-query costs to increase with the explored area. Robot-centric local mapping can bound dense-map memory usage and geometry-intensive computations but discards the global task context needed for persistent, long-horizon exploration.
To overcome this limitation, we present \method{}, a hierarchical spatio-temporal framework that confines dense metric geometry to a robot-centric local map while retaining compact global observation and task context.
The framework organizes exploration context into a cascaded hierarchy spanning observation level and task level. At the \textit{observation level}, lightweight LiDAR-keyframe visibility records preserve fine-grained ray-wise evidence for frontier reasoning. At the \textit{task level}, when an unresolved frontier moves beyond the local-map extent, it is retained as a compact pending task associated with an odometry anchor, rather than retaining its dense local geometry. The task context also includes a sparse topological graph connecting these anchors through retained routes for future revisits.
Upon returning to a selected task region, \method{} progressively recovers the required metric information: the corresponding frontier and exploration viewpoint are reconstructed from current observations, while the topological route is instantiated and validated against the current local map before execution. This hierarchical map representation allows \method{} to explore beyond the spatial extent of its dense local map. As the explored range expands, dense-map memory capacity remains fixed, and exploration-dependent memory growth is confined to compact observation and task context.
The proposed framework achieves 31.98--41.88\% shorter exploration time and a 94--99\% smaller memory footprint of the map representation than the best baseline that completes all three simulated scenes. In real-world deployments, it reduces this footprint by 92.67\% in the forest and 62.42\% at the construction site relative to an explicit point-cloud ikd-tree while preserving long-term task continuity under constrained onboard memory and computation budgets. The project page can be found at \url{https://yizhezhang0418.github.io/flavor.github.io/}
\end{abstract}


\begin{keywords}
LiDAR \sep Large-scale 3D exploration \sep Hierarchical map representation \sep Robot-centric mapping \sep Autonomous UAV
\end{keywords}

\maketitle

\section{Introduction}
\label{sec:introduction}

Autonomous exploration on resource-constrained UAV platforms enables rapid spatial data acquisition for large-scale surveying, infrastructure inspection, and post-disaster assessment, especially in environments where prior maps are incomplete and direct human access is difficult \citep{faria2019autonomous,kerle2019uav}. While human-in-the-loop operation is essential for mission-critical surveying \citep{zhang2026aware}, sustained exploration across vast or hard-to-access environments requires greater onboard autonomy to reduce the need for continuous operator intervention. LiDAR sensing and odometry provide geometric measurements and motion estimates that support surveying and associated decision-making \citep{li2019nrli,zhang2026armor,li2026aeos}. To construct a map autonomously, a UAV must identify unobserved regions, select informative sensing poses, and generate feasible trajectories as the mapped workspace expands \citep{bircher2016receding,zhou2021fuel,zhou2026eame}. Supporting these functions with a persistent global metric map incurs growing memory and query costs.

Existing methods reduce this overhead by making persistent geometry cheaper to store and query \citep{duberg2020ufomap,reijgwart2023efficient}, simplifying planning through multi-resolution and topological abstractions \citep{cao2021tare,chen2022fast,zhang2024falcon}, or retaining selected geometric elements \citep{huang2024efficient,geng2025epic}. Many of these systems still base long-term decisions on metric geometry accumulated over the mission. Fixed-window mapping bounds this geometric workload \citep{ren2024rog}, but discarding old regions also removes the evidence needed to distinguish previously observed boundaries from unfinished tasks. The gap is therefore how to preserve exploration continuity after historical geometry is discarded: the map representation must retain observation context for recognizing prior coverage and task context for recovering unfinished objectives, while confining detailed occupancy, clearance, and collision queries to the current neighborhood.

\begin{figure*}[!t]
  \centering
  \includegraphics[width=\textwidth]{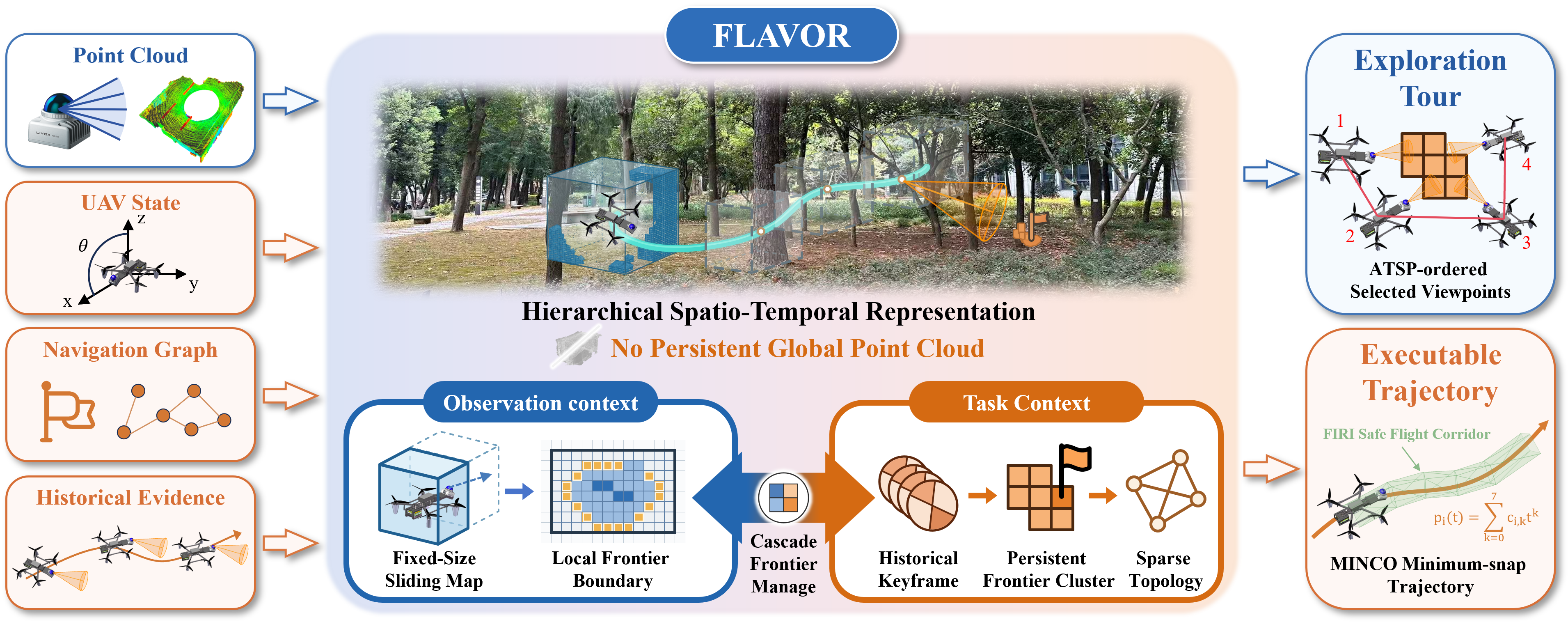}
  \caption{Overview of the \method{} exploration framework. Local geometry and retained mission context are coupled through cascaded frontier updates, preserving exploration continuity without a persistent global point cloud. Selected viewpoints are ordered into an exploration tour. The planner constructs a corridor along the validated local route and optimizes a trajectory within it.}
  \label{fig:framework}
\end{figure*}

To bridge this gap, we propose \method{}, a hierarchical spatio-temporal map representation for large-scale 3D exploration, as illustrated in Fig.~\ref{fig:framework}. By unifying \textbf{F}rontier tasks, a \textbf{L}ocal map, \textbf{A}nchors, historical \textbf{V}isibility, \textbf{O}dometry, and topological \textbf{R}outing, it couples local geometry with retained mission context to support exploration planning without a persistent global point cloud. Our key insight is that historical geometry and historical exploration state support different decisions and therefore need not be retained at the same fidelity. We organize persistent information into two levels: \textit{observation context} retains evidence of prior sensing, while \textit{task context} preserves unfinished objectives together with the anchors and topological routes needed to revisit them. Dense metric geometry remains confined to a robot-centric local map. This separation assigns compact history to long-term exploration decisions and current geometry to viewpoint validation, collision checking, and trajectory generation.

\method{} carries exploration tasks from initial observation through deferred completion in three steps. First, as the UAV observes a region, the observation context preserves visibility evidence for recognizing previously observed boundaries after map shifts. Second, when an unresolved frontier leaves the local window, the task context preserves the pending objective, its associated anchor, and a return route while the surrounding dense geometry is discarded. The task remains pending outside the local map, and its stored anchor and route guide the UAV back for further exploration. Finally, when no valid local target remains, the task context guides a return to a pending region. Current sensing validates each local route segment before execution and, upon re-observation of the task, determines what remains unfinished and supports viewpoint reconstruction. This sequence allows exploration to continue across map shifts while reserving detailed geometric reasoning for the region currently being observed.
The main contributions of this work are summarized as follows:

(1) We propose a hierarchical spatio-temporal map representation that confines dense geometry to a fixed-window local map while retaining compact observation context and task context for large-scale exploration.

(2) We develop cascaded frontier updates that use LiDAR-keyframe visibility to filter repeated boundaries and retain unresolved tasks across map shifts for reconstruction upon revisits.

(3) We develop anchor-associated planning that prioritizes local tasks and retrieves historical tasks through topological routes, with local route validation before execution.

(4) Simulation and real-world experiments demonstrate reduced map memory and computational overhead, along with improved exploration efficiency, compared with the evaluated baselines.

The remainder of this paper is organized as follows. Section~\ref{sec:related_work} reviews exploration-oriented map representations, frontier-based exploration and persistent task management, and hierarchical and topological planning for large-scale 3D exploration. Section~\ref{sec:method} presents the proposed map representation, its online update mechanism, and the anchor-associated exploration framework. Section~\ref{sec:experiments} reports the experimental evaluation, and Section~\ref{sec:discussion} analyzes the resulting design implications and scalability. Finally, Section~\ref{sec:conclusion} concludes the paper.

\section{Related Work}
\label{sec:related_work}

Autonomous exploration couples three functions: maintaining a map, identifying what remains to be observed, and planning motion toward the selected task. In this section, we review map representations, frontier-based exploration and persistent task management, and hierarchical and topological planning.

\subsection{Map Representations for Exploration}

Dense occupancy grids expose unknown--free boundaries directly, but at a fixed resolution their storage is proportional to the voxelized volume. Octree maps reduce repeated storage in homogeneous regions. OctoMap established a probabilistic octree map representation of occupied, free, and implicitly unknown space \citep{hornung2013octomap}; UFOMap made all three states explicit and optimized map queries and updates for robotic planning \citep{duberg2020ufomap}. Wavemap uses a wavelet hierarchy to adapt occupancy-map detail to the spatial structure of the observations \citep{reijgwart2023efficient}. For surface reconstruction, Voxel Hashing allocates voxel blocks around observed surfaces rather than preallocating the full volume \citep{niessner2013real}. Using voxel hashing for storage, Voxblox incrementally derives an ESDF from a TSDF for onboard planning \citep{oleynikova2017voxblox}.

TARE combines dense local data and a detailed local path with sparse global subspaces and a coarse global route; its later dual-resolution formulation makes this local--global division explicit \citep{cao2021tare,cao2023representation}. M3 communicates star-convex free-space polytopes and summarized frontier information \citep{gao2022meeting}. FAOMap stores obstacle and frontier data while omitting extensive free and unknown space \citep{huang2024efficient}. EPIC removes the global occupancy grid by storing surface-observation quality in a spatial hash and constructing an incremental topological graph directly from point clouds \citep{geng2025epic}. Its global point-cloud map nevertheless remains a metric input to planning. Recent work has also focused on local map representations for robotics. ROG-Map keeps occupancy and obstacle inflation in a robot-centric sliding grid, thereby bounding the dense map by a configured local window \citep{ren2024rog}. However, without a global representation, this local map alone offers limited support for large-scale exploration.

In \method{}, occupancy, clearance, and collision queries are restricted to a sliding local map. What persists outside that map is exploration history: observation evidence, unfinished tasks, odometry anchors, and routes previously used to reach them.

\subsection{Frontier-Based Exploration and Persistent Task Management}

Frontier exploration treats the boundary between known free and unknown space as a source of navigation goals \citep{yamauchi1997frontier}. Receding-horizon next-best-view planning instead samples informative trajectories online \citep{bircher2016receding}. Witting et al. record visited locations to bias next-best-view sampling and escape confined-space dead ends \citep{witting2018history}. Dai et al. generate candidate views from frontier voxels organized in an octree and evaluate their information gain by sparse ray casting \citep{dai2020fast}. FUEL incrementally updates a frontier information structure before refining global and local tours \citep{zhou2021fuel}. For surface reconstruction, informed sampling places candidate viewpoints near surface frontiers \citep{kompis2021informed}. FAEP includes flight- and frontier-level factors in global ordering and adapts yaw to observe more frontier cells during transit \citep{zhao2023autonomous}. ERRT integrates information gain, travel distance, and actuation cost in a sampling-based tree to select locally executable next-best trajectories \citep{lindqvist2024tree}. SOAR assigns viewpoints on uncovered surfaces to camera-equipped UAVs \citep{zhang2024soar}. FALCON organizes frontier visitation with a coverage path over incrementally decomposed zones \citep{zhang2024falcon}. EPIC forms exploration boundaries between well- and poorly-observed surface patches \citep{geng2025epic}.

These approaches use frontiers and viewpoints to guide exploration. With a sliding local map, however, a frontier may disappear because it leaves the window rather than because its task is complete. Previously observed boundaries can also appear new when discarded regions re-enter the map as unknown. \method{} therefore uses LiDAR-keyframe visibility records to filter repeated boundaries and persistent frontier cells to retain unresolved tasks.

\subsection{Hierarchical and Topological Exploration Planning}

Early GBPlanner formulations couple a local rapidly exploring random graph with a sparse global graph that returns a robot from dead ends to previously identified exploration boundaries \citep{dang2019graph}. Graph-based topological exploration similarly groups traversable regions and plans over their connectivity \citep{yang2021graph}. SphereMap fills free space with intersecting spheres and converts them into a multi-layer planning graph \citep{musil2022spheremap}, whereas sparse-skeleton methods obtain nodes and edges from the structure of observed free space \citep{chen2022fast}. A persistent rewiring tree can refine informative paths over repeated planning iterations \citep{schmid2020efficient}.

Historical connectivity does not guarantee that a route remains collision-free. \method{} associates tasks with odometry anchors and validates the local route against the current map before execution. FIRI constructs a convex free-space corridor around the validated route \citep{wang2025fast}. MINCO provides the polynomial trajectory parameterization, and GCOPTER optimizes the parameterized trajectory within the corridor under the required geometric and dynamic constraints \citep{wang2022geometrically}.

\section{Method}
\label{sec:method}

\begin{figure*}[!h]
  \centering
  \includegraphics[width=\textwidth]{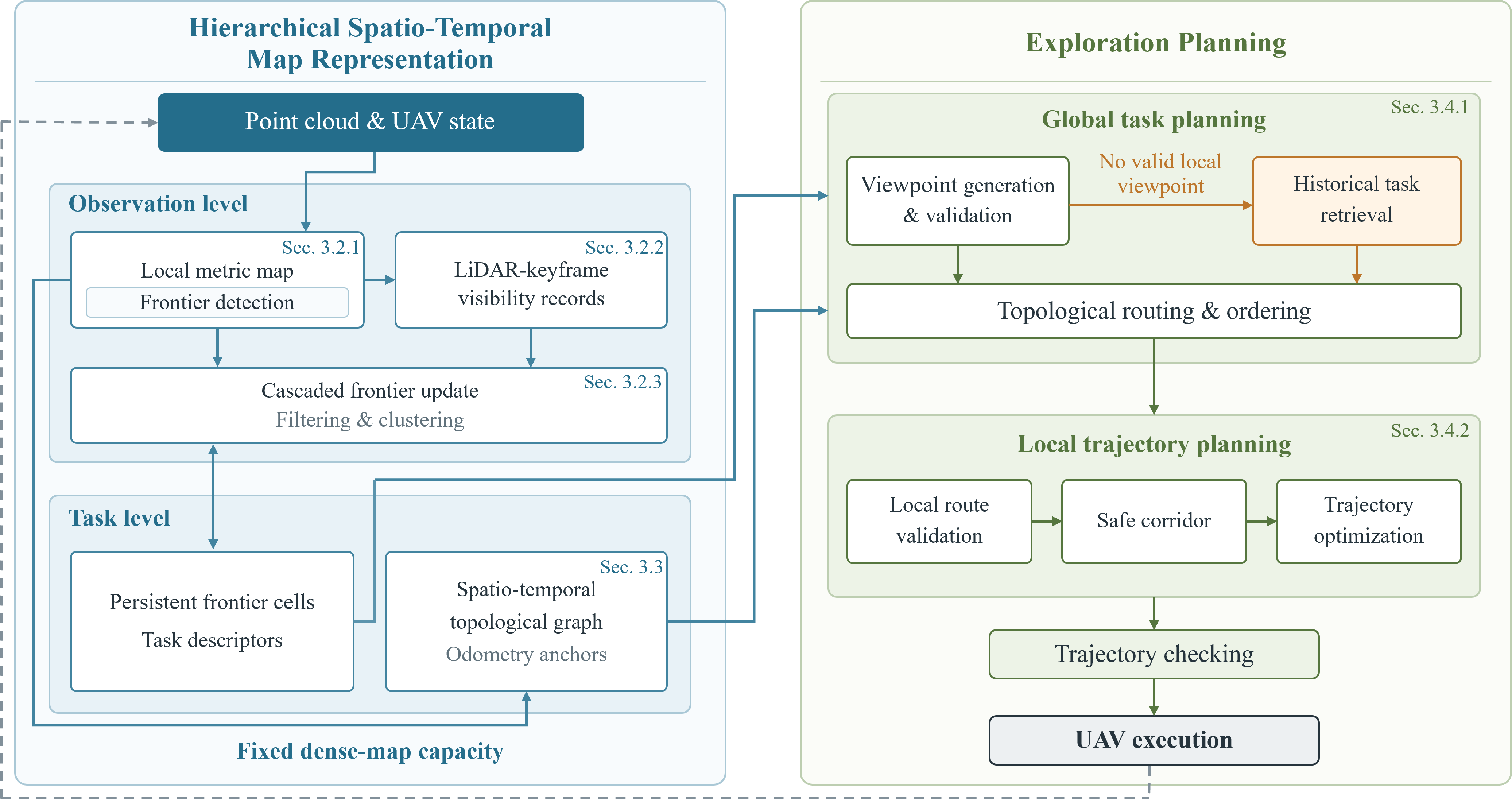}
  \caption{Overall workflow of \method{}. The hierarchical spatio-temporal map representation maintains local geometry, observation context, and task context for exploration planning. Local route validation and trajectory checking precede execution. Section numbers refer to the corresponding method descriptions.}
  \label{fig:method_overview}
\end{figure*}

\method{} integrates a hierarchical spatio-temporal map representation with exploration planning, as illustrated in Fig.~\ref{fig:method_overview}. At the \textit{observation level}, a robot-centric local metric map supports frontier extraction, while LiDAR-keyframe visibility records filter previously observed boundaries. Cascaded frontier updates preserve unresolved tasks across map shifts and refresh them using current observations. At the \textit{task level}, persistent frontier cells and task descriptors retain unfinished tasks, while odometry anchors and the spatio-temporal topological graph guide their recovery beyond the local window. Planning prioritizes valid local viewpoints and retrieves historical tasks when none remain. Local trajectory planning validates the local route, constructs a safe corridor, and optimizes the trajectory using the MINCO representation. The trajectory is checked before UAV execution.

\subsection{Problem Formulation and Hierarchical Map Representation}
\label{sec:problem}

\subsubsection{Problem Formulation}

We consider a UAV exploring a bounded three-dimensional workspace $\mathcal W\subset\mathbb R^3$ using onboard LiDAR. At each planning cycle, it selects an informative sensing pose and a dynamically feasible trajectory. A fixed-size robot-centric local map bounds the dense mapping cost, while task context retains unfinished objectives and their return routes across map shifts without maintaining a globally queryable point cloud or occupancy field.

At time $t$, an external localization module provides the vehicle state
\begin{equation}
  \mathbf{x}_t=({}^{W}\mathbf{p}_{B,t},{}^{W}\mathbf{q}_{B,t},{}^{W}\mathbf{v}_{B,t}).
\label{eq:planner_input}
\end{equation}
The module also provides a registered point cloud ${}^{W}\mathcal P_t$ in the fixed world frame $\mathcal F_W$. Here, ${}^{W}\mathbf{p}_{B,t}$, ${}^{W}\mathbf{q}_{B,t}$, and ${}^{W}\mathbf{v}_{B,t}$ denote the UAV position, orientation, and velocity, respectively. Table~\ref{tab:frames} summarizes the frames, and ${}^{A}\mathbf{R}_B$ denotes rotation from $\mathcal F_B$ to $\mathcal F_A$. Persistent frontier cells, task attributes, graph routes, and trajectories are expressed in $\mathcal F_W$, while field-of-view (FoV) sector assignment uses the instantaneous LiDAR frame $\mathcal F_{L_t}$.

\begin{table}[]
\centering
\caption{Coordinate frames and map convention.}
\label{tab:frames}
\small
\renewcommand{\tabularxcolumn}[1]{m{#1}}
\begin{tabularx}{\columnwidth}{>{\raggedright\arraybackslash}m{1.18cm}Y Y}
\toprule
Symbol & Definition & Primary use \\
\midrule
$\mathcal F_W$ & Fixed localization frame & Persistent state, graph routes, and trajectories \\
\addlinespace[3pt]
$\mathcal F_{B_t}$ & Moving UAV body frame & Vehicle state and dynamics \\
\addlinespace[3pt]
$\mathcal F_{L_t}$ & Moving LiDAR frame & Field of view and visibility rays \\
\addlinespace[3pt]
$\mathcal M_t^L$ & World-aligned sliding grid & Occupancy, distance, and frontier queries \\
\bottomrule
\end{tabularx}
\end{table}

The LiDAR orientation is given by ${}^{W}\mathbf{R}_{L,t}={}^{W}\mathbf{R}_{B,t}\mathbf{R}_y(\theta_L)$, where $\theta_L$ is the fixed sensor pitch. We project a registered point into the LiDAR frame as
\begin{equation}
{}^{L_t}\mathbf{p}=
({}^{W}\mathbf{R}_{L,t})^{\mathsf T}
({}^{W}\mathbf{p}-{}^{W}\mathbf{p}_{L,t}),
\label{eq:world_lidar_transform}
\end{equation}
where ${}^{W}\mathbf{p}_{L,t}$ defines the origin of the visibility rays. The map window follows the UAV while its grid axes remain aligned with $\mathcal F_W$. Equation~\eqref{eq:world_lidar_transform} assigns points to angular sectors in the LiDAR frame. Each sector's representative direction is then expressed in $\mathcal F_W$ for ray traversal and comparison with the local occupancy map.

\subsubsection{Hierarchical Spatio-Temporal Map Representation}

To retain exploration history without extending the local metric map, the hierarchical spatio-temporal map representation combines current local geometry with persistent observation context and task context. At time $t$, its state is
\begin{equation}
\mathcal S_t=\left(\mathcal M_t^L,\mathcal H_t,\mathcal G_t\right),
\qquad
\mathcal H_t=\left(\mathcal D_t,\mathcal F_t^G,\mathcal C_t^G,\mathcal A_t\right),
\label{eq:system_state}
\end{equation}
where $\mathcal M_t^L$ is the robot-centric metric map, $\mathcal H_t$ the retained exploration state, and $\mathcal G_t$ a sparse topological graph. LiDAR-keyframe visibility records $\mathcal D_t$ provide observation context. Persistent frontier cells $\mathcal F_t^G$, historical cluster descriptors $\mathcal C_t^G$, odometry anchors $\mathcal A_t$, and $\mathcal G_t$ provide task context. The superscript $G$ denotes persistence in the world frame, not a global metric grid. The local window fixes dense-map capacity, while the storage of compact observation context and task context scales with retained mission history.

Frontier information is maintained at both cell and cluster levels. Local cluster geometry and candidate viewpoints are evaluated against current map evidence, while unresolved cells persist in $\mathcal F_t^G$. For mission-level scheduling, each stored cluster has a compact task descriptor $\eta_j\in\mathcal C_t^G$:
\begin{equation}
\eta_j=\left(j,\boldsymbol{\mu}_j,\mathbf{n}_j,
[\mathbf{b}_j^-,\mathbf{b}_j^+],N_j,\hat{\mathbf{v}}_j,\hat\psi_j,a_j\right),
\label{eq:task_descriptor}
\end{equation}
where $j$ is the cluster identifier, $\boldsymbol{\mu}_j$ and $\mathbf{n}_j$ are its center and mean normal, $[\mathbf{b}_j^-,\mathbf{b}_j^+]$ is its axis-aligned bounding box, and $N_j$ is the active-cell count. The associated anchor $a_j\in\mathcal A_t$ has position $\mathbf{p}_{a_j}$ and yaw $\psi_{a_j}$. The cached scheduling pose $(\hat{\mathbf{v}}_j,\hat\psi_j)$ stores the last valid sensing pose when available. Otherwise, it is initialized as $(\boldsymbol{\mu}_j,\psi_{a_j})$. This pose ranks historical tasks, while current measurements reconstruct the frontier and validate an executable sensing pose after the UAV returns.

Free-space connectivity is represented by
\begin{equation}
\mathcal G_t=(\mathcal V_t,\mathcal E_t,\omega_t,\Pi_t),
\label{eq:topological_state}
\end{equation}
where $\mathcal V_t$ and $\mathcal E_t$ are the node and edge sets, $\omega_t(e)$ is edge traversal cost, and $\Pi_t(e)$ is its route polyline. The graph connects sparse free-space nodes, historical odometry anchors, the current odometry node, and transient task or viewpoint nodes. It preserves routes established by local obstacle-clearance checks and recorded odometry, while $\mathcal H_t$ retains observation evidence and unfinished tasks. Occupancy and collision queries use the current local metric map.

\subsection{Robot-Centric Mapping and Cascaded Frontier Representation}

\subsubsection{Local Metric Map and Frontier Extraction}
\label{sec:local_map}

To bound dense-map storage and support frontier extraction, we maintain a fixed-size robot-centric voxel map~\citep{ren2024rog} that stores occupancy, inflated obstacles, and distance-to-obstacle information. Let $h_j$ denote the array half-width along axis $j$ in cells, $\mathbf{h}=(h_x,h_y,h_z)$, and $r$ the base resolution. The number of cells along each axis and total capacity are
\begin{equation}
  n_j=2h_j+1,
  \qquad N_L=n_x n_y n_z.
\label{eq:local_map_size}
\end{equation}
For fixed $\mathbf{h}$ and $r$, the dense-map capacity is $O(N_L)$ and is independent of the distance traveled by the UAV.

\begin{figure}[]
  \centering
  \includegraphics[width=\columnwidth]{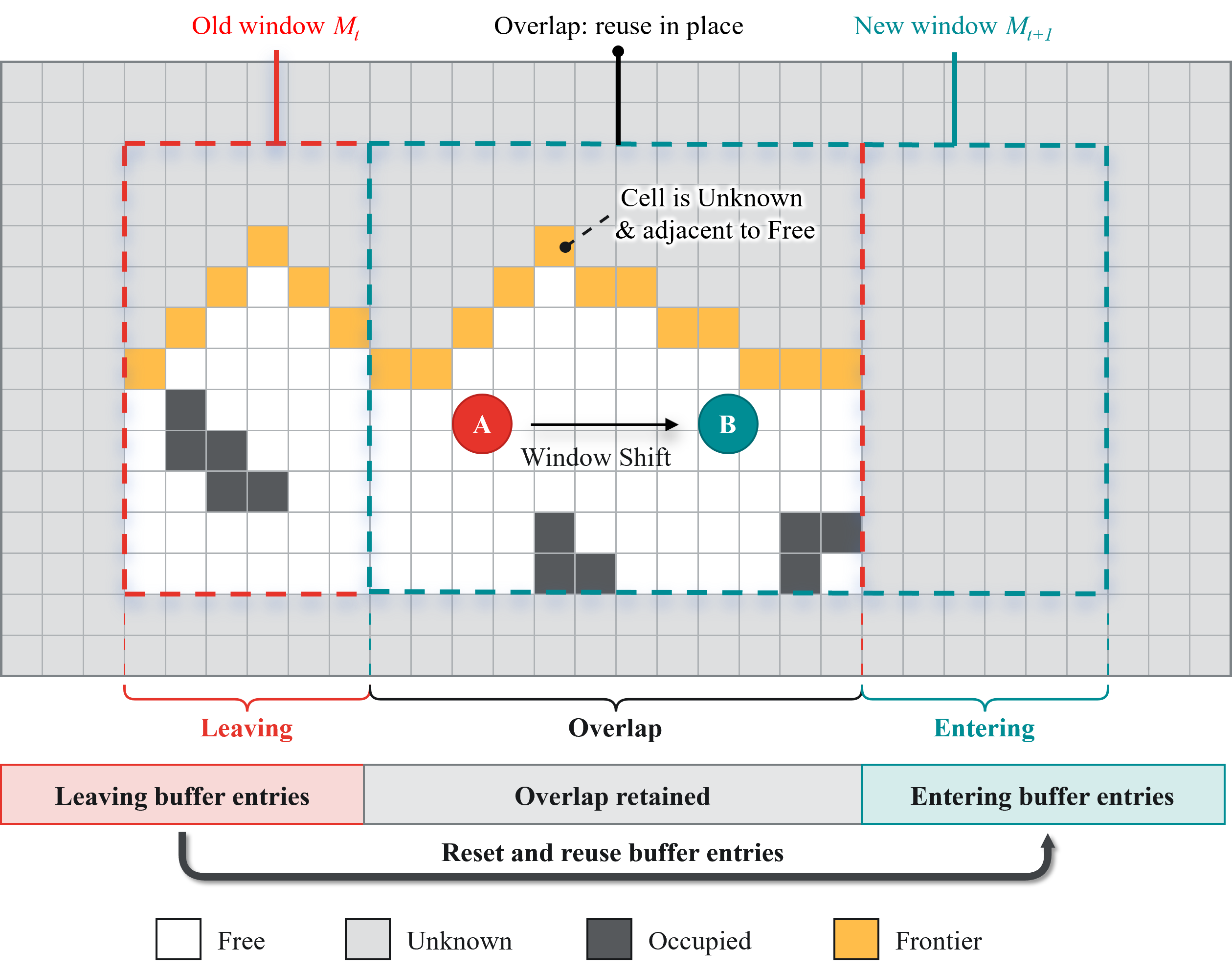}
  \caption{Sliding local map and frontier extraction in two dimensions. The overlapping map region is retained, and leaving storage is reused for entering cells. Yellow cells are raw frontiers; white, light gray, and dark gray indicate free, unknown, and occupied cells.}
  \label{fig:sliding_map}
\end{figure}

The local grid uses a three-dimensional circular buffer with modular indexing. The window recenters when the UAV's displacement from its current center exceeds a prescribed threshold. As illustrated in Fig.~\ref{fig:sliding_map}, the overlap $\mathcal B_t=\mathcal M_t^L\cap\mathcal M_{t+1}^L$ is preserved at existing memory addresses, while leaving entries are reset and reused for entering cells, initialized as unknown.

Registered LiDAR scans update occupancy within the current window by raycasting, including overlap cells receiving new observations. Hit and miss evidence is fused using clamped log-odds updates. Occupancy changes incrementally update the obstacle-inflation layer, while local obstacle distances support clearance queries for viewpoint validation and trajectory planning.

Raw frontiers are unknown cells adjacent to known free space:
\begin{equation}
\begin{aligned}
\mathcal F_t^{L,\mathrm{raw}}=\{\mathbf{c}\mid {}&m_t(\mathbf{c})=\text{unknown},\\
&\exists\mathbf{c}'\in\mathcal N_{26}(\mathbf{c}):m_t(\mathbf{c}')=\text{free}\},
\end{aligned}
\label{eq:local_frontier}
\end{equation}
where $m_t$ denotes occupancy state and $\mathcal N_{26}(\mathbf{c})$ is the 26-connected neighborhood. Candidates pass current-map and historical-visibility checks before spatial clustering. As observations change, local clusters may grow, split, merge, disappear, or be reconstructed after revisits. We consequently do not assume a permanent one-to-one correspondence between a local cluster identifier and a physical boundary.

\subsubsection{LiDAR-Keyframe Visibility Records}
\label{sec:visibility}

We introduce LiDAR-keyframe visibility records to retain observation context and reject repeated frontiers when previously observed regions re-enter the sliding window. Each selected keyframe stores a compact angular visibility descriptor:
\begin{equation}
\begin{aligned}
d_i&=(\mathbf{p}_i,\mathbf{q}_i,\mathcal Z_i),\\
\mathcal Z_i&=(\mathbf{z}_{i,1},\ldots,\mathbf{z}_{i,B}),\\
\mathbf{z}_{i,b}&=(f_{i,b},o^-_{i,b},o^+_{i,b},g_{i,b}).
\end{aligned}
\label{eq:descriptor}
\end{equation}
Here, $(\mathbf{p}_i,\mathbf{q}_i)$ is the LiDAR pose and $B=B_{\mathrm{az}}B_{\mathrm{el}}$ is the number of azimuth--elevation sectors. In each sector, $f_{i,b}$ ends the observed-free prefix, $o^-_{i,b}$ and $o^+_{i,b}$ bound the occupied surface interval, and $g_{i,b}$ counts guard-band voxels immediately behind it, all encoded in voxel steps. A zero tuple indicates that no usable interval is retained. The first valid scan initializes the records. A later descriptor is retained only after a minimum time interval and sufficient translation or rotation.
Visibility records and odometry anchors use independent sampling criteria: the former retain angular observation evidence, while the latter provide return locations for historical tasks.

\begin{figure}[]
  \centering
  \includegraphics[width=\columnwidth]{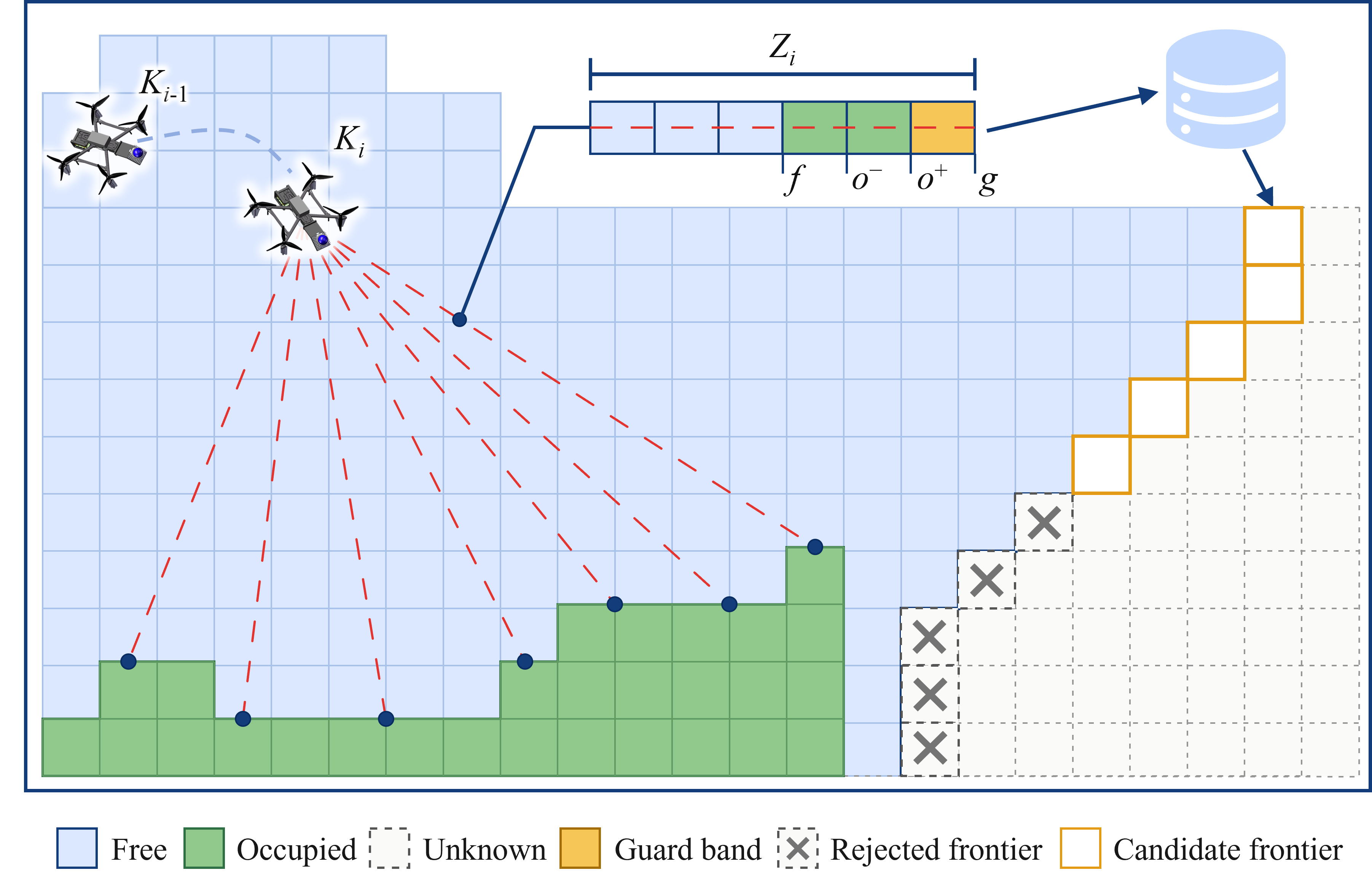}
  \caption{LiDAR-keyframe visibility and frontier filtering. The inset shows one sector's voxel-step encoding of free space (blue), an occupied surface (green), and a guard band (yellow). Historical visibility rejects repeated frontiers (gray crosses), leaving unexplained candidates (orange outlines).}
  \label{fig:keyframe_visibility}
\end{figure}

For each FoV sector, the nearest return establishes the range supported by the current scan. Traversing the local map outward from the sensor along the representative ray yields the observed-free prefix and occupied surface interval, as illustrated in Fig.~\ref{fig:keyframe_visibility}. The guard band continues along that ray behind the surface until the next obstacle, configured voxel-count limit, sensing limit, or map boundary, whichever occurs first. It supports frontier filtering without assigning free or occupied states. These intervals summarize visibility without retaining every intersected voxel. Under the bounded sensing horizon, each field occupies one byte, so the angular component requires $4B$ bytes per descriptor, excluding pose and indexing metadata.

During a frontier update, a spatial index selects descriptors whose sensing domains overlap the affected region. Their representative rays are replayed over candidate cells. A candidate is removed if directly covered by a stored interval or if any of its 26 neighbors has historical free or occupied support. Neighboring guard-band coverage alone does not trigger removal. The surviving cells enter the cascaded frontier update, without reconstructing a global occupancy field.

\subsubsection{Cascaded Frontier Update}
\label{sec:task_memory}

The cascaded frontier update combines current observations with observation context to retain unfinished frontier cells as task context across map shifts. We associate observations of the same frontier cell using a discrete key defined by its position $\mathbf{p}$ in the fixed world frame:
\begin{equation}
  \mathbf{k}(\mathbf{p})=
  \left\lfloor\frac{\mathbf{p}-\mathbf{b}_{\min}}{r_f}\right\rfloor,
\label{eq:frontier_key}
\end{equation}
where $r_f$ is the frontier resolution and $\mathbf{b}_{\min}$ is the workspace lower bound. Each cell carries geometric attributes (position and normal), temporal evidence (observation count and lifecycle state), and cluster and anchor associations. A newly detected cell remains provisional until it receives sufficient observation support. Once activated, it persists across map shifts until current evidence resolves the boundary or its lifecycle criterion is reached.

Local observations update the persistent frontier cells in $\mathcal F_t^G$, which in turn support the recovery of local clusters after map shifts. Reconstruction is limited to clusters overlapping the affected region. Within that region, newly detected frontiers and active persistent cells are reclustered with existing cluster cells supported by current frontier evidence or active unresolved task memory. The resulting clusters provide updated geometry, cached viewpoints, and anchor associations for the task descriptors.

On a revisit, new observations reveal which persistent cells within the local window still belong to a frontier. These cells form a new local cluster, while resolved cells are retired. Cells without sufficient evidence remain undecided, including those outside the window. If no frontier is locally supported but undecided cells remain, their mean position guides further observation. An executable sensing pose is then regenerated and validated from the new observations.

\subsection{Spatio-Temporal Topological Graph}
\label{sec:topology}

Within task context, we introduce the spatio-temporal topological graph $\mathcal G_t$ to preserve route connectivity beyond the local metric window by coupling a regional spatial skeleton with persistent odometry anchors. Current local geometry updates spatial connections, while recorded UAV odometry maintains a temporal backbone across map shifts. Its vertex set comprises spatial skeleton vertices $\mathcal V_t^S$, historical anchor vertices $\mathcal V_t^A$, the current odometry vertex $o$, and temporary query vertices $\mathcal V_t^Q$,
\begin{equation}
  \mathcal V_t=\mathcal V_t^S\cup\mathcal V_t^A\cup\{o\}\cup\mathcal V_t^Q.
\label{eq:topo_vertex_types}
\end{equation}
Each anchor in $\mathcal A_t$ identifies a persistent vertex in $\mathcal V_t^A$. Edge polylines retain routes after their supporting geometry leaves the map. The initial route segment within the local map is validated against current geometry before trajectory generation.

\subsubsection{Regional Skeleton Construction}

To represent local free-space connectivity for route planning, we partition the workspace into fixed, world-aligned cuboids with side lengths $\boldsymbol{\ell}=(\ell_x,\ell_y,\ell_z)$. A position $\mathbf p\in\mathcal W$ has region index
\begin{equation}
  \iota_j(\mathbf p)=
  \left\lfloor\frac{p_j-b_{\min,j}}{\ell_j}\right\rfloor,
  \qquad j\in\{x,y,z\},
\label{eq:topo_region_index}
\end{equation}
where $\mathbf b_{\min}$ is the workspace lower bound. Regions overlapping the exploration domain are indexed by a spatial hash, with each region retaining its skeleton vertices and an anchor reference when available. To cover both observed surfaces and the free-space passages leading to them, candidate regions are identified from range-limited LiDAR returns and samples along segments from the sensor to the corresponding region centers. At each cycle, the skeleton vertices in a limited number of candidate regions are reconstructed from $\mathcal M_t^L$, with priority given to nearby regions, while the remaining regions retain their vertices.

Within selected regions, recursive sphere coverage~\citep{geng2025epic} extracts representative free-space vertices from spheres placed at subregion centers, with radii equal to the local obstacle distance. Subregions whose sphere radii fall below the minimum are bisected along axes that permit further subdivision; for accepted spheres, recursion continues outside their inscribed cubes. Recursion stops once coverage is satisfied or minimum subregion dimensions are reached. Spheres are grouped when their center distance is less than the sum of their radii minus an overlap margin. Each group is represented by a clearance-qualified sphere center near the region center, or by the largest-radius sphere center when no such candidate is available. Multiple vertices can thus represent disconnected free space within one region. The spheres are temporary, while the regional vertices and their graph connections form the retained spatial structure.

After a regional update, reconstructed skeleton vertices are matched to existing vertices by location. Matched vertices retain their identities and connections for edge revalidation, while previous vertices without a match are removed together with their incident edges and stored paths. New vertices are added to their corresponding regions.

\subsubsection{Incremental Edge Maintenance}

To maintain spatial connectivity as local observations change, new vertices are connected to candidates in the same or six face-adjacent regions. A straight segment is accepted if it satisfies local clearance; otherwise, A* searches the endpoints' axis-aligned bounding box expanded by $\ell_j/2$ on each side along axis $j$. A valid path forms a bidirectional edge with polyline $\Pi_t(e)=(\mathbf p_0,\ldots,\mathbf p_m)$ and weight
\begin{equation}
  \omega_t(e)=
  \sum_{k=0}^{m-1}\|\mathbf p_{k+1}-\mathbf p_k\|_2.
\label{eq:topo_edge_weight}
\end{equation}
Retained spatial paths are rechecked and shortened against the updated local map. Invalid paths are repaired by a new search or removed if repair fails, and weights are updated according to the resulting polylines. Unconnected neighboring pairs are also tested.

\subsubsection{Odometry Anchors and Temporal Connectivity}

\begin{figure*}[]
  \centering
  \includegraphics[width=.98\textwidth]{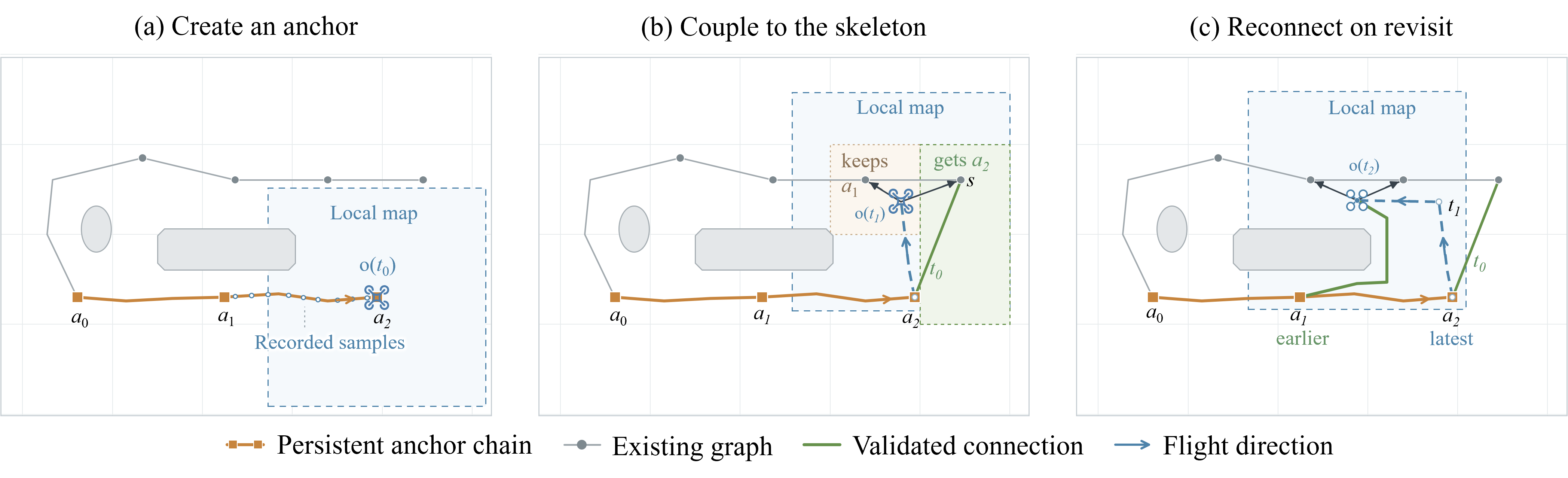}
  \caption{Odometry anchors along a continuous flight, with current odometry vertex $o(t_i)$. (a) Inserting $a_2$ stores the recorded route from $a_1$. (b) A later region update connects $a_2$ to skeleton vertex $s$, preserving existing region--anchor associations. (c) The current odometry vertex gains a validated shortcut to $a_1$ while retaining its recorded connection to $a_2$. Dashed boxes denote the local map, and arrows indicate flight direction. Anchor-related edges are bidirectional. Only relevant connections are shown.}
  \label{fig:odometry_anchor}
\end{figure*}

Odometry anchors preserve temporal connectivity for revisiting historical tasks as the local map shifts. Each anchor is a persistent graph vertex at a recorded UAV position, identified by its historical odometry index. Anchors are spaced to keep the temporal backbone sparse during revisits. A new anchor is inserted in a registered region when clearance is sufficient and its Euclidean distance from every existing anchor exceeds $\delta_a$. The first anchor is initialized once the region and clearance conditions are met.

As illustrated in Fig.~\ref{fig:odometry_anchor}, recorded odometry connects successive anchors through persistent bidirectional edges. Each edge retains the sampled flight route, shortened where local clearance checks permit. The anchors and their routes remain available throughout the mission, preserving return paths after the surrounding metric geometry leaves the local map. Persistent anchor identities provide stable references for historical tasks as the spatial skeleton is reconstructed from new observations.

To connect this temporal backbone to the spatial skeleton, a new anchor is associated with its own region and neighboring regions sharing a face, edge, or corner, provided they have no existing anchor association. During subsequent regional updates, collision-free paths between the anchor and skeleton vertices establish bidirectional edges. These connections follow the same checking, repair, and deletion rules as spatial edges. Graph route searches start from the current odometry vertex $o$, which represents the UAV position and yaw. Its outgoing connections to nearby skeleton vertices are rebuilt using locally checked paths, while accumulated odometry maintains a bidirectional connection to the latest anchor. Collision-free paths to earlier anchors within $\delta_a$ provide additional bidirectional shortcuts during revisits.

Historical task descriptors reference anchors as return locations, and multiple tasks may share an anchor. Temporary viewpoint vertices connect through locally checked paths for graph queries and are removed with their edges afterward. Concatenating the stored edge polylines yields a route through skeleton vertices, anchors, or both. The shortest-path distance $d_{\mathcal G_t}(u,v)$ is computed using the edge weights in Eq.~\eqref{eq:topo_edge_weight}.

\subsection{Exploration Planning}
\label{sec:scheduling}

Building on the frontier tasks and spatio-temporal topological graph, exploration planning combines current local observation evidence with retained mission tasks. Global task planning selects and orders sensing targets, while local trajectory planning converts the selected graph route into dynamically feasible motion using current local geometry.

\subsubsection{Global Task Planning}

Global task planning follows a local-first cycle to explore current frontiers and recover unfinished historical tasks. Current local observations support viewpoint generation, validation, and ordering for local clusters. When no valid local viewpoint remains, the planner retrieves historical tasks with reachable anchors. Their prefiltering cost combines the route to the anchor with the estimated remaining approach distance:
\begin{equation}
  \hat J(o,i)=d_{\mathcal G_t}(o,a_i)
  +\|\mathbf{p}_{a_i}-\hat{\mathbf{v}}_i\|_2.
\label{eq:memory_prefilter}
\end{equation}
Here, $a_i$ and $\hat{\mathbf{v}}_i$ are the anchor and cached scheduling position of task $\eta_i$ defined in Eq.~\eqref{eq:task_descriptor}. The graph distance measures route length through current skeleton and anchor connections, including shortcuts, while the Euclidean term estimates the remaining approach distance.

\begin{figure}[]
  \centering
  \includegraphics[width=\columnwidth]{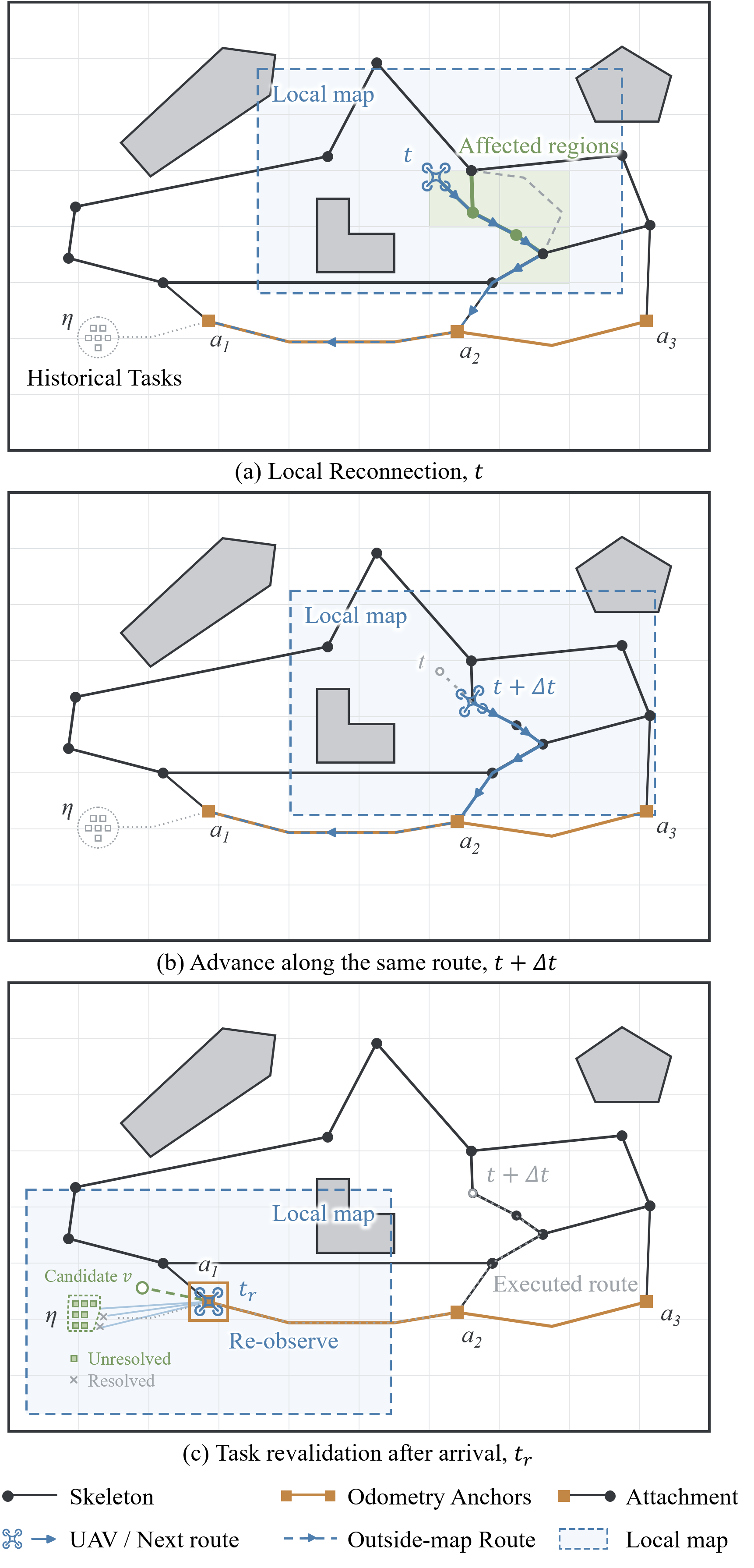}
  \caption{Historical-task retrieval. (a) Regional skeleton updates provide a route to task anchor $a_1$. (b) The UAV follows the route as its local map shifts. (c) Re-observation separates unresolved frontier cells (green) from resolved cells (crosses), and regenerated viewpoint $v$ is evaluated for reachability.}
  \label{fig:spatio_temporal_topology}
\end{figure}

As illustrated in Fig.~\ref{fig:spatio_temporal_topology}, the UAV follows a graph route toward the anchor and uses new observations to reconstruct unresolved clusters through the frontier update in Section~\ref{sec:task_memory}. Sensing targets are then regenerated and validated from the recovered clusters.

For both current and recovered clusters, candidate viewpoints are assessed against the current point cloud and graph for obstacle clearance, flight-region limits, and reachability. For viewpoints outside the local map window, where dense clearance evidence is unavailable, traversability is instead inferred from the skeletonized region in the topological graph, and the viewpoint is revalidated against new observations after the UAV approaches it. Sensing distance, viewing direction, line of sight, and LiDAR FoV determine valid frontier coverage. We select the yaw that maximizes coverage at each candidate position and choose the viewpoint with the highest coverage among reachable candidates with valid coverage.

To account for both route length and altitude variation when ordering tasks, we evaluate a graph route $\pi=(\mathbf{x}_0,\ldots,\mathbf{x}_m)$ between $u$ and $v$, with $m\geq1$, using the travel-time surrogate
\begin{equation}
\begin{aligned}
T_{\mathcal G}(u,v)=\frac{2}{v_{\max}}
\sum_{k=0}^{m-1}\big(&\|\mathbf{x}_{k+1}-\mathbf{x}_k\|_2\\
&+0.5\left|(\mathbf{x}_{k+1})_z-(\mathbf{x}_k)_z\right|\big).
\end{aligned}
\label{eq:topo_cost}
\end{equation}
The first term measures route length, while the second penalizes altitude variation. Validated viewpoint vertices serve as endpoints for local tasks. Historical tasks use their anchors as route endpoints, with $2\|\mathbf p_{a_i}-\hat{\mathbf v}_i\|_2/v_{\max}$ estimating the remaining approach time. The cost between two historical tasks is the sum of the route cost between their anchors and the estimated travel time from each anchor to its task's cached scheduling position. These estimates guide task ordering, while local trajectory optimization determines the executable motion.

For multiple candidates, we use A* search on $\mathcal G_t$ to compute pairwise routes and construct the cost matrix from Eq.~\eqref{eq:topo_cost} and the applicable endpoint approach terms. We formulate visit ordering from $o$ as an asymmetric traveling-salesman problem (ATSP) and obtain a heuristic tour using LKH~\citep{helsgaun2000effective}.

\subsubsection{Local Trajectory Planning}

Local trajectory planning converts the selected graph route into dynamically feasible motion using current local geometry. During replanning, the current reference position on the executing trajectory is prepended to the selected graph route. The resulting route is subdivided into sufficiently close waypoints, and its initial segment within the local map is checked against current geometry and repaired before trajectory generation. Initial position, velocity, acceleration, and jerk are taken from the executing trajectory to ensure a smooth transition. Occupied points from the local ROG-Map support corridor construction.

The position trajectory $\mathbf{p}(t)=[x(t),y(t),z(t)]^\top$ and yaw trajectory $\psi(t)$ use the piecewise-polynomial MINCO representation. FIRI~\citep{wang2025fast} constructs convex regions around the local route; a connected sequence containing the initial state forms the flight corridor. GCOPTER~\citep{wang2022geometrically} jointly optimizes the spatial and temporal parameters of $\mathbf{p}(t)$ within it under a flight-time cost and velocity, body-rate, tilt, and thrust limits. The yaw trajectory is optimized separately to follow the forward flight direction and reach the desired final yaw.

After optimization, the position trajectory is checked against the point cloud using a hard clearance margin, and its velocity is checked against the configured limit. An imminent collision or velocity violation causes the result to be rejected. During execution, synchronized LiDAR--odometry updates integrate new observations into the local map and recheck the remaining trajectory. A newly detected collision triggers replanning, and execution stops if collision is too close.
\section{Experiments}
\label{sec:experiments}

\subsection{Evaluation Metrics and Baselines}
\label{sec:setup}

We evaluate \method{} using exploration time, map memory footprint, average flight speed, and planning latency. Baselines are TARE~\citep{cao2021tare}, ERRT~\citep{lindqvist2024tree}, FALCON~\citep{zhang2024falcon}, and EPIC~\citep{geng2025epic}. Estimates are reported in decimal megabytes (MB).

\begin{figure}[]
  \centering
  \includegraphics[width=\columnwidth]{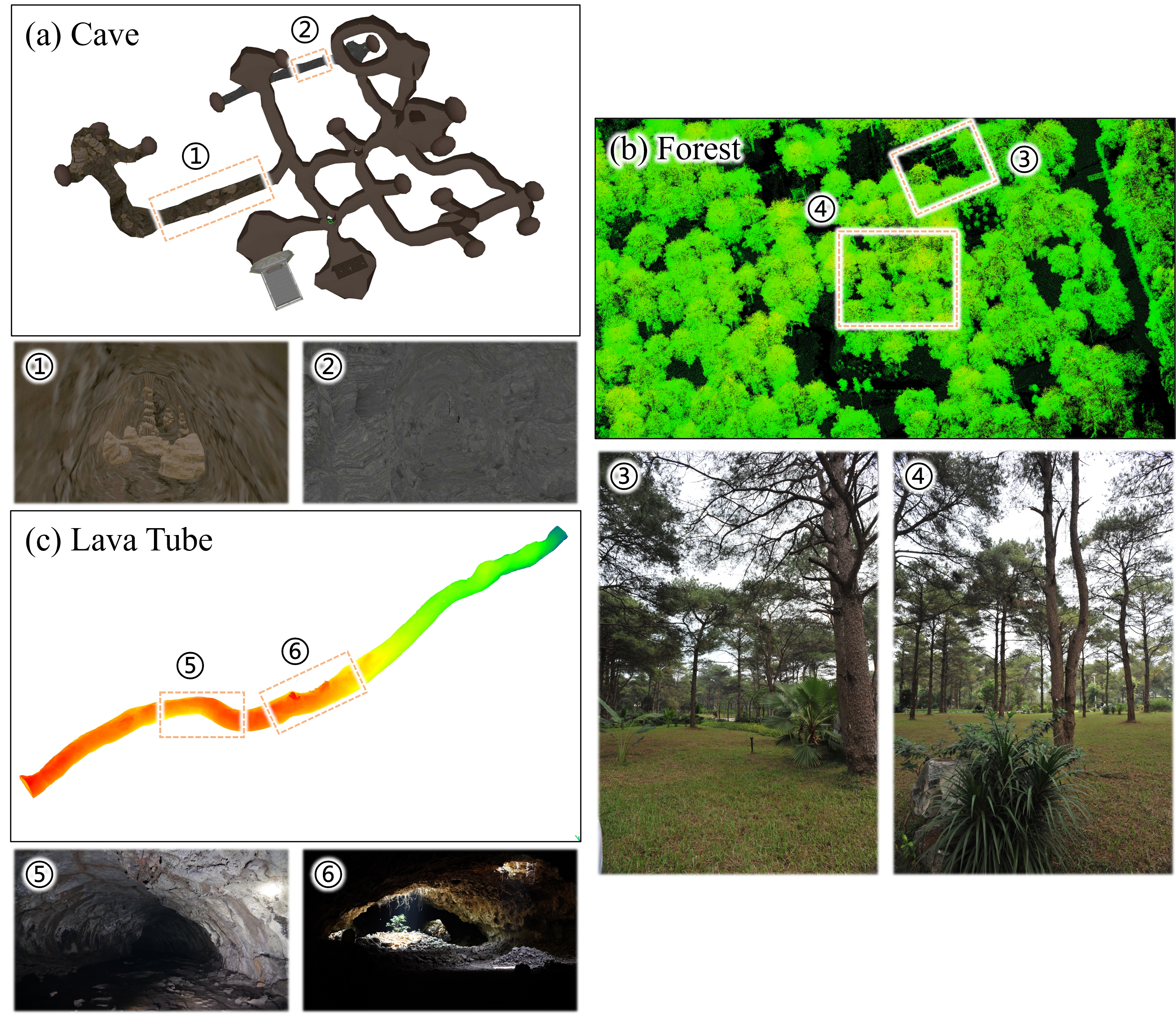}
  \caption{Simulation scenes: (a) Cave, (b) Forest, and (c) Lava Tube. Dashed boxes and matching numbers locate the rendered views or on-site photographs.}
  \label{fig:simulation_scene}
\end{figure}

Tests use three large-scale point-cloud scenes, as illustrated in Fig.~\ref{fig:simulation_scene}: Cave [$224 \times 330 \times 41$]~m$^3$, Forest [$108 \times 88 \times 5$]~m$^3$, and Lava Tube [$313 \times 62 \times 25$]~m$^3$. Cave uses the DARPA SubT Cave World dataset~\citep{koval2020subterranean}. Forest is reconstructed from terrestrial laser scanner surveys collected in Guangxi, China, registered following Dong et al.~\citep{dong2020registration}. Lava Tube is located in Hainan, China. These scenes pose distinct exploration challenges: interconnected tunnels and vertical transitions in Cave, vegetation-induced occlusions in Forest, and restricted clearance along the elongated passage in Lava Tube.

All planners are evaluated in MARSIM~\citep{kong2023marsim} using identical flight dynamics and point-cloud rendering procedures, with a maximum planning speed of 4.0~m/s. All simulation experiments are conducted on a computer equipped with an Intel Core i7-14700KF processor, an NVIDIA GeForce RTX 4090 GPU with 24~GB of memory, and 32~GB of system memory. Coverage progression and map memory are recorded in the same instrumented trial for each method-scene pair. Table~\ref{tab:simulation_config} lists the key configuration parameters of \method{}.

\begin{table*}[]
  \centering
  \renewcommand{\arraystretch}{1.1}
  \caption{Simulation configuration of \method{}.}
  \label{tab:simulation_config}
  \begin{tabular*}{\linewidth}{@{\extracolsep{\fill}} l c c c @{}}
    \toprule
    \textbf{Scene} & \textbf{Local-map window (m)} & \textbf{Map resolution (m)} & \textbf{Visibility resolution/range} \\
    \midrule
    Cave      & $40\times40\times36$ & 0.4 & $2^\circ$, 0.1--16 m \\
    Forest    & $36\times36\times10$ & 0.4 & $2^\circ$, 0.1--18 m \\
    Lava Tube & $48\times48\times16$ & 0.4 & $2^\circ$, 0.1--19 m \\
    \bottomrule
  \end{tabular*}
\end{table*}

\subsection{Evaluation in Simulation}
\label{sec:simulation_results}

\subsubsection{Exploration Efficiency}
\label{sec:exploration_efficiency}

\begin{figure*}[]
  \centering
  \includegraphics[width=.98\textwidth]{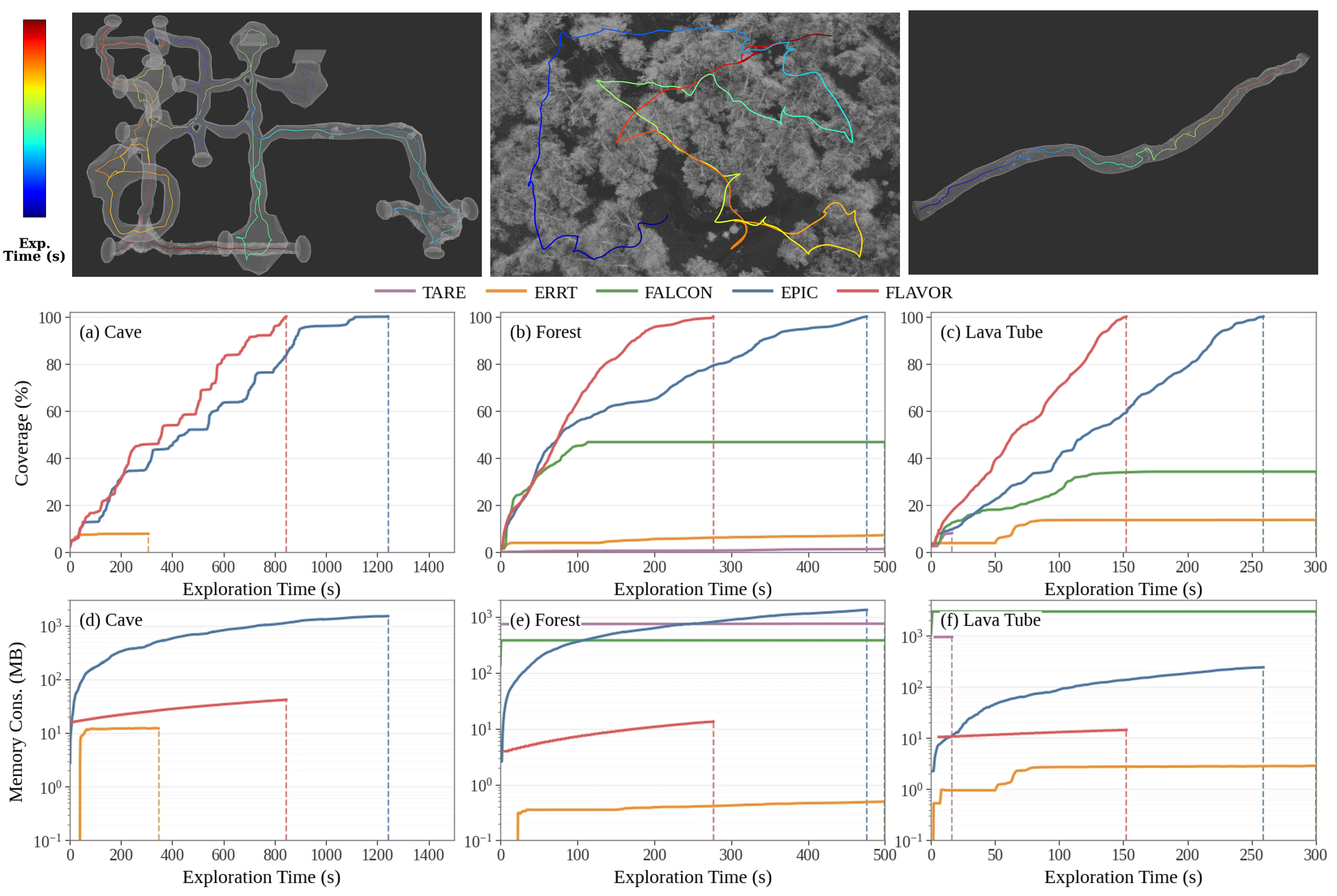}
  \caption{Exploration trajectories, coverage progress, and the memory footprint of the map representation in the point-cloud scenes simulated with MARSIM. The columns show Cave, Forest, and Lava Tube from left to right. The top row shows the trajectory of \method{} colored by elapsed exploration time, the middle row shows normalized surface coverage, and the bottom row shows the memory footprint of each planner's map representation on a logarithmic vertical scale.}
  \label{fig:simulation_benchmark_progress}
\end{figure*}

In Cave, TARE and FALCON fail to initialize and produce no valid runs, leaving ERRT, EPIC, and \method{} for comparison. Fig.~\ref{fig:simulation_benchmark_progress} shows the executed trajectories, normalized coverage progression, and map memory footprint, with all five planners included in Forest and Lava Tube. EPIC is the only baseline that completes all three scenes and therefore serves as the reference for completion-time comparisons.

The coverage curves show sustained progress through scene completion, with especially clear separation from EPIC in Forest and Lava Tube. This behavior is consistent with the cascaded frontier and anchor-associated planning framework: persistent frontiers retain unfinished objectives beyond the local window, anchors and routes guide the UAV back, and current observations support viewpoint reconstruction. Completion of scenes larger than the local-map dimensions in Table~\ref{tab:simulation_config} demonstrates that the framework sustains exploration beyond its dense geometric window.

As reported in Table~\ref{tab:overall_efficiency}, \method{} completes all three scenes with the shortest completion times. Relative to EPIC, exploration time decreases by 31.98\%, 41.88\%, and 41.26\% in Cave, Forest, and Lava Tube, respectively. Average velocity, measured along the integrated three-dimensional odometry trajectory, is 16.52--65.53\% higher than EPIC's across these environments. 

Together with the shorter completion times, these gains highlight \method{}'s ability to sustain fast, mission-scale exploration while keeping dense geometric storage confined to a fixed local window.

\begin{table*}[]
  \centering
  \renewcommand{\arraystretch}{1.1}
  \caption{Mission-level exploration performance in the point-cloud scenes simulated with MARSIM. Bold indicates the best completion result in each scene.}
  \label{tab:overall_efficiency}
  \small
  \begin{tabular}{llrrc}
    \toprule
    \textbf{Scene} & \textbf{Method} & \textbf{Completion Time (s)} & \textbf{Avg. Vel. (m/s)} & \textbf{Coverage Reached} \\
    \midrule
    \multirow{5}{*}{Cave}
      & TARE             & --              & --            & No  \\
      & ERRT             & --              & 1.08          & No  \\
      & FALCON           & --              & --            & No  \\
      & EPIC             & 1242.50         & 2.80          & Yes \\
      & \textbf{\method{}} & \textbf{845.20} & \textbf{3.26} & \textbf{Yes} \\
    \midrule
    \multirow{5}{*}{Forest}
      & TARE             & --              & 2.06          & No  \\
      & ERRT             & --              & 0.39          & No  \\
      & FALCON           & --              & 0.17          & No  \\
      & EPIC             & 476.40          & 1.84          & Yes \\
      & \textbf{\method{}} & \textbf{276.87} & \textbf{3.04} & \textbf{Yes} \\
    \midrule
    \multirow{5}{*}{Lava Tube}
      & TARE             & --              & 1.68          & No  \\
      & ERRT             & --              & 0.57          & No  \\
      & FALCON           & --              & 0.41          & No  \\
      & EPIC             & 259.27          & 1.71          & Yes \\
      & \textbf{\method{}} & \textbf{152.30} & \textbf{2.40} & \textbf{Yes} \\
    \bottomrule
  \end{tabular}
  \\[2pt]
  \parbox{0.88\textwidth}{\footnotesize Completion time is reported at scene completion. The Cave velocity comparison includes ERRT, EPIC, and \method{}.}
\end{table*}

\subsubsection{Representation and Planning Efficiency}
\label{sec:system_efficiency}

\method{} retains visibility evidence, unresolved frontier tasks, odometry anchors, and sparse topological routes across local-map shifts. Dense occupancy and clearance information remains within a robot-centric ROG window of fixed dimensions, supporting frontier revalidation, viewpoint construction, local route repair, collision checking, and trajectory generation. Thus, dense geometric storage is bounded while compact mission history persists.

The memory curves in Fig.~\ref{fig:simulation_benchmark_progress} show modest growth for \method{}, whereas EPIC's accumulated point-cloud state increases to substantially higher values as exploration proceeds. Table~\ref{tab:representation_efficiency} reports the final map memory footprints for these two methods, both of which complete all three scenes, allowing comparison at scene completion.

\begin{table}[]
  \centering
  \renewcommand{\arraystretch}{1.1}
  \caption{Memory footprint of the map representation. Reduction is relative to EPIC, and bold indicates the result of \method{}.}
  \label{tab:representation_efficiency}
  \small
  \begin{tabular}{lrrr}
    \toprule
    \textbf{Scene} & \textbf{EPIC (MB)} & \textbf{\method{} (MB)} & \textbf{Reduction} \\
    \midrule
    Cave      & 1510.69 & \textbf{41.93} & \textbf{97.22\%} \\
    Forest    & 1341.21 & \textbf{13.48} & \textbf{99.00\%} \\
    Lava Tube & 242.89  & \textbf{14.56} & \textbf{94.00\%} \\
    \bottomrule
  \end{tabular}
\end{table}

\method{} reduces the map memory footprint by 94.00--99.00\% relative to EPIC. The representations differ in that EPIC retains an expanding global point-cloud ikd-tree, whereas \method{} bounds dense geometry within the local ROG map. Outside that map, compact persistent records preserve the information needed to recover unfinished objectives and resume exploration without retaining the corresponding raw geometry. The measured memory advantage thus accompanies completed exploration in all three scenes.

\begin{figure}[]
  \centering
  \includegraphics[width=\columnwidth]{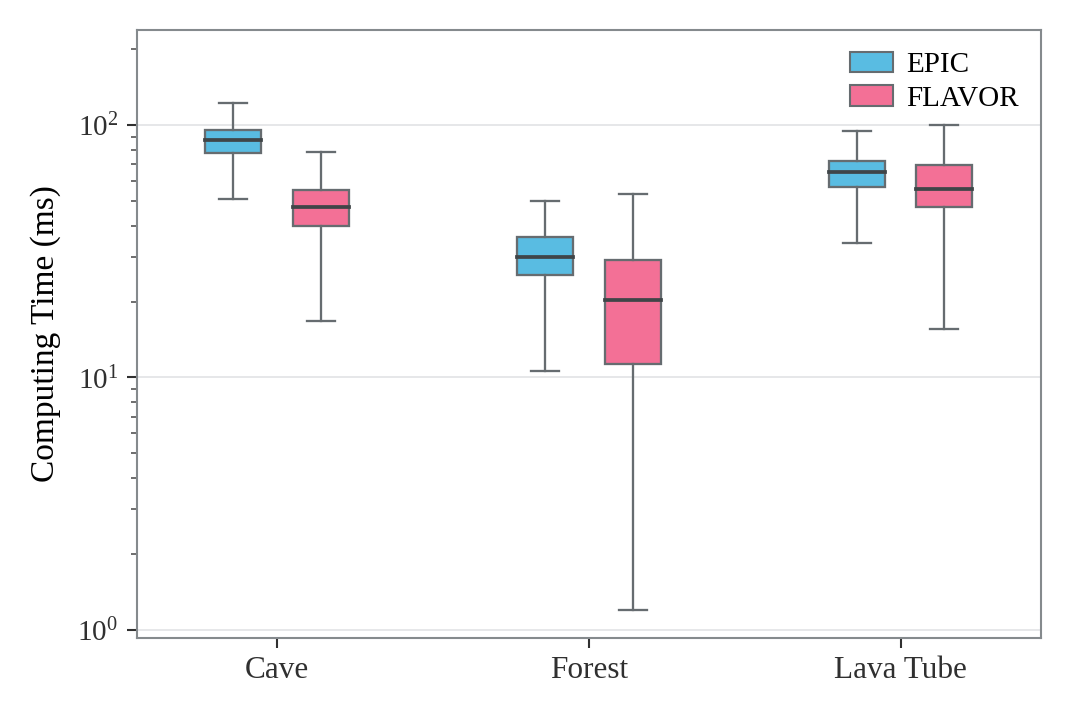}
  \caption{Planning latency of EPIC and \method{} in Cave, Forest, and Lava Tube. Boxes show the interquartile range, and horizontal lines mark the medians. The vertical axis is logarithmic.}
  \label{fig:runtime_comparison}
\end{figure}

\begin{table}[]
  \centering
  \renewcommand{\arraystretch}{1.1}
  \caption{Mean $\pm$ standard deviation of planning latency. Bold indicates the lower mean latency.}
  \label{tab:planning_efficiency}
  \small
  \begin{tabular}{lrrr}
    \toprule
    \textbf{Scene} & \textbf{EPIC (ms)} & \textbf{\method{} (ms)} & \textbf{Reduction} \\
    \midrule
    Cave      & $85.76 \pm 16.52$ & $\mathbf{47.94 \pm 11.73}$ & \textbf{44.10\%} \\
    Forest    & $30.36 \pm 8.23$  & $\mathbf{21.43 \pm 11.59}$ & \textbf{29.42\%} \\
    Lava Tube & $64.61 \pm 13.24$ & $\mathbf{59.50 \pm 17.64}$ & \textbf{7.92\%} \\
    \bottomrule
  \end{tabular}
\end{table}

Fig.~\ref{fig:runtime_comparison} and Table~\ref{tab:planning_efficiency} compare planning latency with EPIC. Median planning latency is lower for \method{} in all three environments. Mean latency reductions are 44.10\% in Cave, 29.42\% in Forest, and 7.92\% in Lava Tube. Relative to EPIC, the standard deviation is lower in Cave but higher in Forest and Lava Tube, indicating greater variability across planning cycles in the latter scenes. The lower mean latency is consistent with the task--anchor--route abstraction, which focuses topology maintenance and candidate ordering on compact task and connectivity records. Retained routes guide returns to regions where current observations reconstruct executable tasks.

\subsection{Evaluation in Real-World Deployment}
\label{sec:real_world_experiments}

\begin{figure}[]
  \centering
  \includegraphics[width=.98\columnwidth]{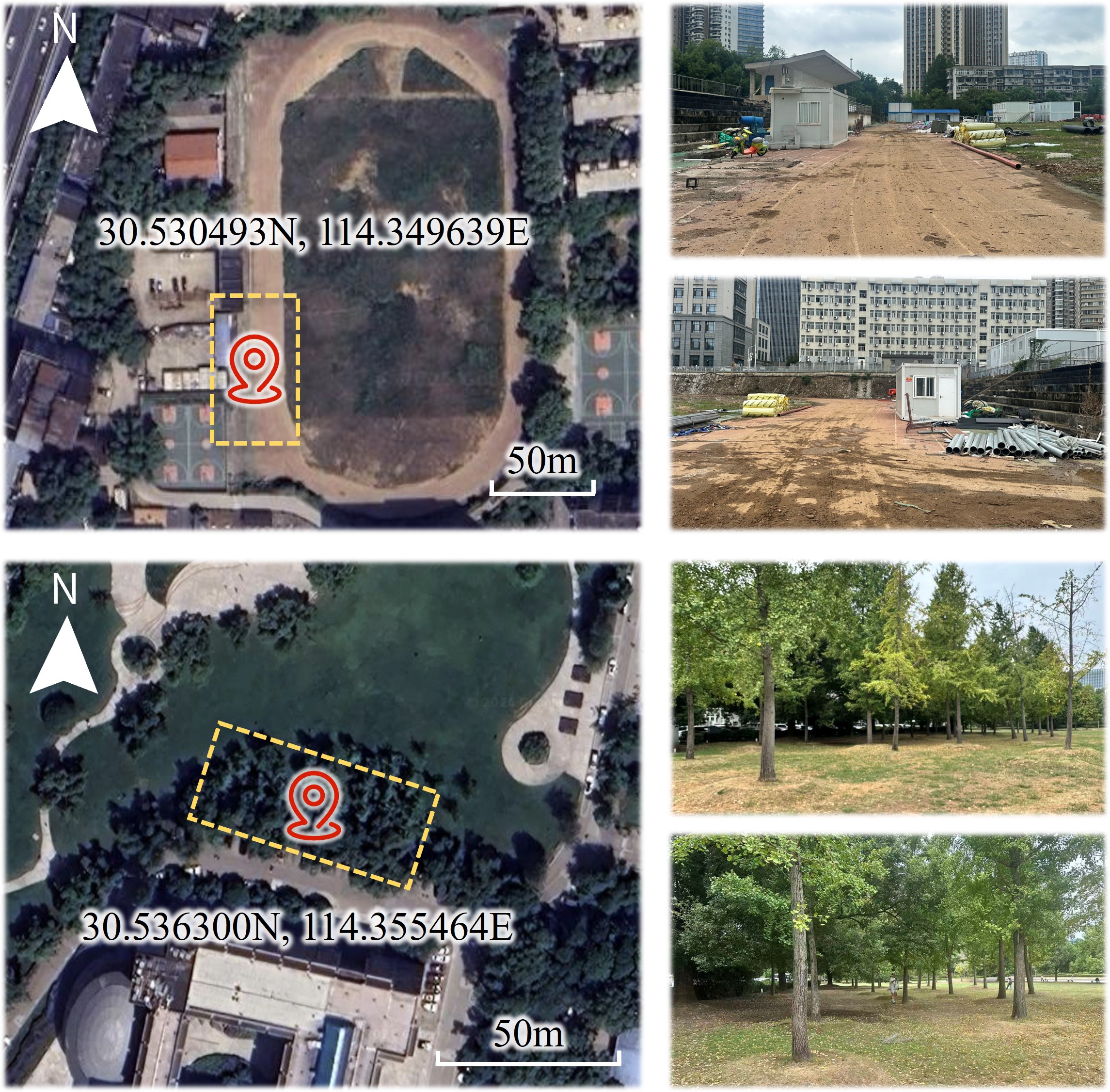}
  \caption{Real-world experiment sites: construction site (top) and forest (bottom). Satellite views on the left show the geographic coordinates and experiment areas marked by yellow dashed boxes. Photographs on the right show the conditions at each site.}
  \label{fig:real_world_scene}
\end{figure}

We deployed \method{} in a forest and a construction site, as shown in Fig.~\ref{fig:real_world_scene}. The forest contains trees with overlapping branches and foliage, while the construction site contains temporary buildings and scattered construction materials.

\begin{figure}[]
  \centering
  \includegraphics[width=.98\columnwidth]{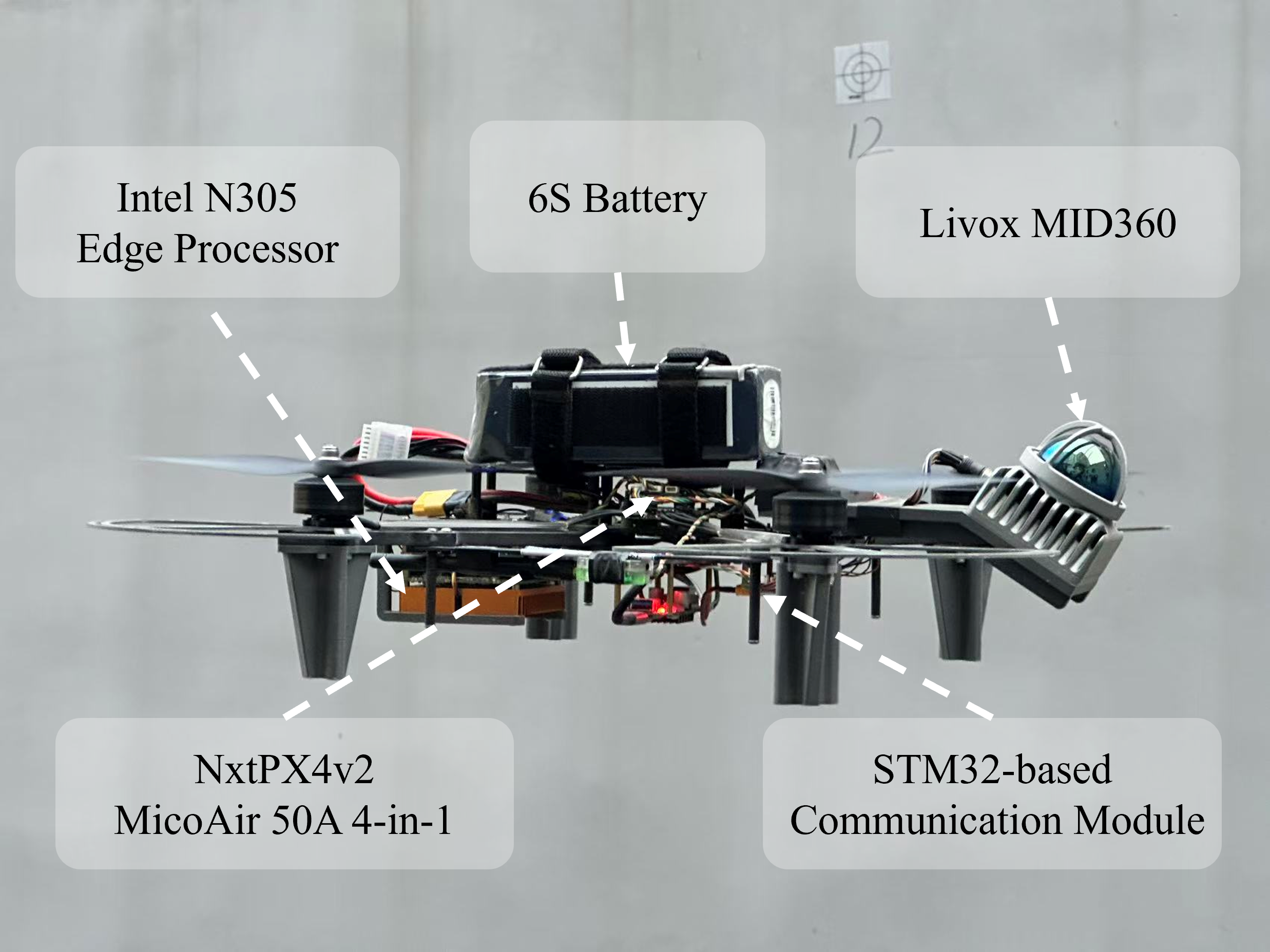}
  \caption{UAV platform used in the real-world deployments, showing the Intel N305 edge computer, Livox MID360 LiDAR, NxtPX4v2 flight controller, and STM32-based communication module.}
  \label{fig:real_world_uav_platform}
\end{figure}

Our custom-built UAV, shown in Fig.~\ref{fig:real_world_uav_platform}, carries an x86 Intel N305 edge computer and a Livox MID360 LiDAR, which provides a $360^\circ$ horizontal FoV and a vertical FoV from $-7^\circ$ to $52^\circ$. A custom-built STM32-based module transfers LiDAR data to the onboard computer. The LiDAR was mounted at a fixed pitch angle of $-30^\circ$, and a modified FAST-LIO2 provided onboard state estimation~\citep{xu2022fast}. For safety, the maximum velocity was set to $1.2$~m/s for both deployments. 

The forest and construction-site exploration regions measured approximately [$60\times27\times3$]~m$^3$ and [$50\times27\times3.5$]~m$^3$, respectively. The robot-centered ROG map was configured as [$24\times24\times10$]~m$^3$ in the forest and [$36\times36\times10$]~m$^3$ at the construction site, both at $0.4$~m resolution. In each environment, the mission extent exceeded the corresponding local-map width, requiring the dense map window to move with the UAV throughout exploration.

\begin{figure*}[!b]
  \centering
  \includegraphics[width=0.98\textwidth]{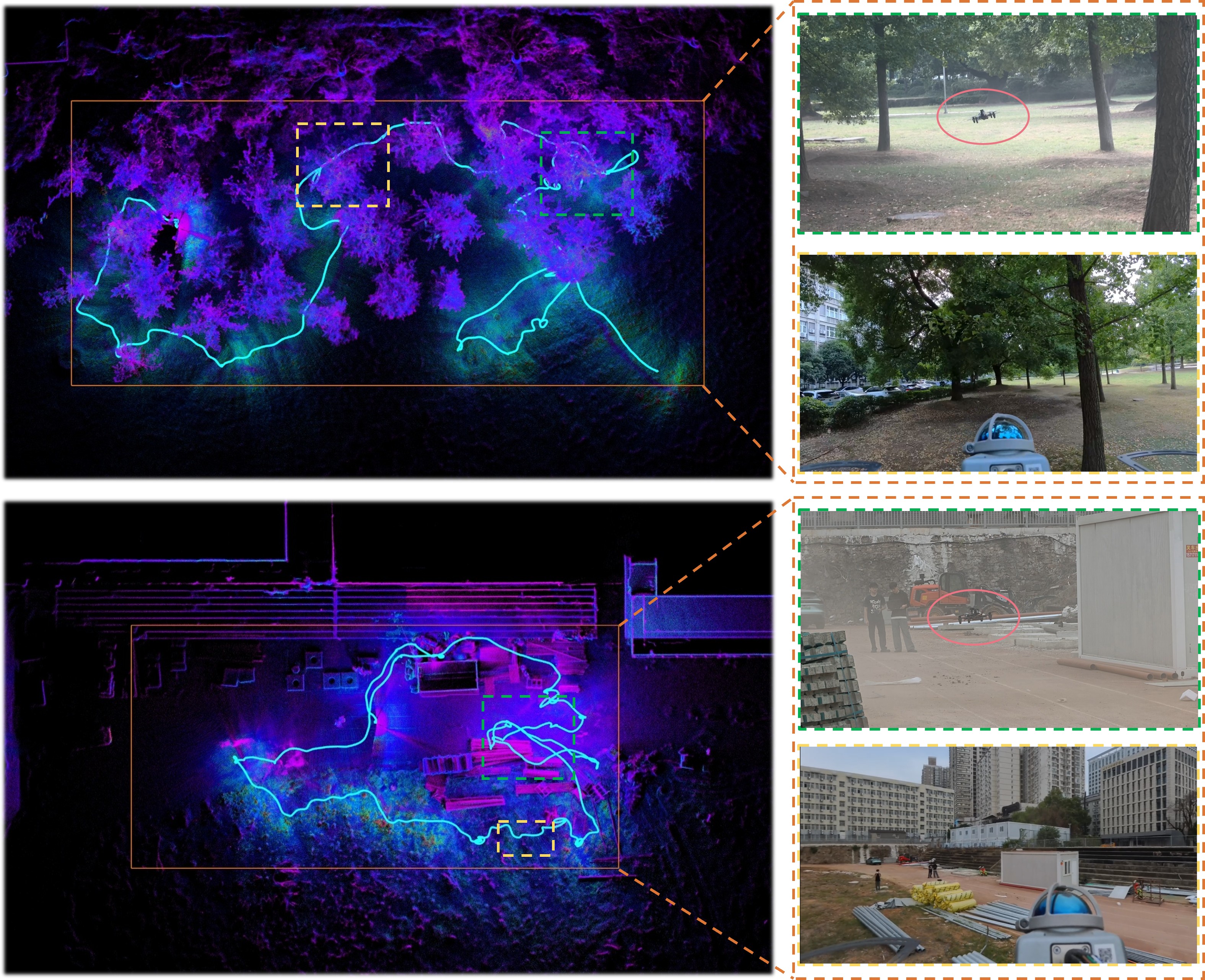}
  \caption{Offline visualization of the real-world forest (top) and construction-site (bottom) deployments. The left panels overlay the executed UAV trajectory (cyan) and the top-down exploration bounding box (orange) on the reconstructed point-cloud map. Dashed boxes mark selected trajectory locations, and the snapshots on the right show the corresponding third-person and onboard first-person views. The global point clouds are shown only for visualization and were not used by the online planner.}
  \label{fig:real_world_flight_scenes}
\end{figure*}

The complete exploration pipeline ran on the onboard processor throughout both missions. Fig.~\ref{fig:real_world_flight_scenes} shows the executed trajectories and selected flight views. The global point clouds were used only for offline visualization and were not queried by the online planner. The UAV traversed $248.18$~m in $318.89$~s in the forest and $259.22$~m in $318.74$~s at the construction site, with average flight speeds of $0.778$~m/s and $0.813$~m/s, respectively. During autonomous exploration, dense occupancy and clearance queries remained confined to the robot-centered map. As reported in Table~\ref{tab:real_world_representation_efficiency}, \method{} reduces the map memory footprint by $92.67\%$ in the forest and $62.42\%$ at the construction site relative to an explicit point-cloud ikd-tree. These savings come from confining dense geometry to the local ROG map and retaining compact visibility records for exploration history. The larger relative saving in the forest also reflects its smaller local-map window and the larger footprint of the reference ikd-tree.

\begin{table}[t]
  \centering
  \renewcommand{\arraystretch}{1.1}
  \setlength{\tabcolsep}{3pt}
  \caption{Memory footprint of the map representation in real-world deployment. The measured components of \method{} are the historical visibility representation and local ROG buffers. Reduction is relative to the explicit point-cloud ikd-tree, and bold indicates the result of \method{}.}
  \label{tab:real_world_representation_efficiency}
  \small
  \begin{tabular}{lccc}
    \toprule
    \textbf{Scene} & \textbf{ikd-tree (MB)} & \textbf{\method{} (MB)} & \textbf{Reduction} \\
    \midrule
    Forest            & 50.64 & \textbf{3.71} & \textbf{92.67\%} \\
    Construction Site & 19.80 & \textbf{7.44} & \textbf{62.42\%} \\
    \bottomrule
  \end{tabular}
\end{table}

In the forest, foliage and thin branches produced sparse, nonuniform returns and frequent occlusions. \method{} filters local free--unknown boundaries using current occupancy and retained visibility evidence, reducing dependence on point-cloud surface criteria~\citep{geng2025epic}. At the construction site, dust raised by rotor wash produced transient LiDAR returns that occasionally caused abrupt stops when treated as obstacles. The sliding local map discards affected geometry once it leaves the window, limiting the persistence of false obstacles without explicitly detecting or tracking dynamic objects.

These deployments demonstrate exploration beyond the local-map extent using onboard computation. The local ROG map supplies geometry for frontier revalidation, collision checking, and viewpoint generation. Historical visibility and frontier states retain observation and task context as the window moves. When an unresolved region re-enters the map, the planner uses this context and current observations to resume exploration without retaining dense geometry for the entire environment, thus allowing the UAV to complete exploration across the full mission area.

\section{Discussion}
\label{sec:discussion}

\subsection{How the Components Work Together}

When geometry outside the local window is discarded, \method{} retains observation context and task context. The observation context uses historical visibility records to filter frontier responses when previously mapped regions re-enter the window. The task context retains unresolved objectives through persistent frontier cells and task descriptors, together with anchors and routes for later recovery. These two forms of context support different decisions: visibility evidence helps identify repeated boundaries, while task state records where exploration remains unfinished. Together, they preserve the context needed to distinguish repeated observations from pending tasks after local geometry has been discarded.

Anchor-associated planning connects these persistent tasks to regions where they can be observed and evaluated again. When no valid local viewpoint remains, the selected task's anchor provides a return location, and the topological graph supplies route guidance. As the UAV returns, current observations support frontier reassessment and viewpoint reconstruction within the local map. If the task remains outside the window, an intermediate approach brings it closer to the sensing region. The local route is checked and repaired before trajectory generation. Retained task records and routes guide the UAV back to unfinished regions, while the current local map is used to check the local route and trajectory before flight.

\subsection{Sustaining Fast Exploration with Compact Task Memory}

\method{} sustains exploration efficiency by decoupling local geometric reasoning from mission-scale task management. EPIC constructs and validates its topology through geometric queries on accumulated point clouds~\citep{geng2025epic}. In \method{}, occupancy and clearance queries use a fixed robot-centric map, while compact task records and stored routes support long-range decisions. Affected-region updates and local-first scheduling further concentrate computation on exploration tasks within the local map. This division reduces the work required for each planning update and allows guidance routes to respond promptly to new observations, supporting continuous trajectory execution as exploration advances.

Task persistence also promotes more continuous motion. EPIC's frontiers separate well- and poorly-observed surfaces, so residual patches can attract repeated visits to satisfy sensing-distance and incidence-angle requirements~\citep{geng2025epic}. This mechanism is consistent with the back-and-forth motion observed in its runs. In confined spaces, surface-quality refinement can also favor wall-adjacent viewpoints, consuming clearance beyond the enforced safety threshold. In \method{}, local-first scheduling with a previous-goal reference reduces unnecessary target changes, while the coupled history mechanisms preserve access to unfinished tasks across map shifts. Current observations then support viewpoint reconstruction and local-route validation.

TARE, FALCON, and ERRT differ in how they retain map information and plan exploration. TARE retains global surface data and orders active subspaces~\citep{cao2021tare}; FALCON maintains voxel-derived zones and computes their global coverage tour~\citep{zhang2024falcon}. Their storage and ordering workloads remain tied to represented geometry and task count. ERRT uses a local sampling volume without an integrated strategy to recover unexplored regions beyond its reach~\citep{lindqvist2024tree}. \method{} combines bounded local geometry with compact mission-wide task recovery, preserving access to unfinished objectives beyond the local planning horizon.

\subsection{Scaling Mission Context with Bounded Local Geometry}
\label{sec:method_efficiency}

Volumetric and surface-based methods retain geometric information over volumes and observed surfaces, respectively. A dense voxel map at resolution $r$ requires $O(V/r^3)$ storage for mapped volume $V$. Approximately uniform surface samples at spacing $r$ require $O(A/r^2)$ storage for observed surface area $A$. In \method{}, the occupancy, inflation, and distance arrays require $O(N_L)$ storage, with $N_L$ fixed by the local window and resolution. The circular buffer supports $O(1)$ voxel access and retains overlapping cells in place as the local window moves~\citep{ren2024rog}. Exploration beyond this window adds compact observation context and task context, including visibility descriptors, unresolved frontier cells, task descriptors, odometry anchors, and stored topological routes. In particular, $N_D$ visibility descriptors with $B$ angular bins require $O(N_D B)$ storage. Each descriptor has an angular payload of $4B$ bytes, with its pose and spatial-index metadata stored separately. Frontier and route storage grows with the retained cells, task records, graph elements, and sparse route points. As exploration expands, memory grows with observation context and task context, while dense-map storage remains fixed.

This division confines frequent geometric operations to the fixed local map. Occupancy integration costs $O(P_tL)$ for $P_t$ scan points and at most $L$ voxel steps per ray. A full local frontier pass costs $O(N_L)$. Visibility-descriptor construction costs $O(P_t+BL)$, processing the scan once and traversing at most $L$ voxel steps along each of the $B$ representative rays. Historical visibility queries select descriptors whose sensing domains overlap affected regions, where frontier clustering and topology updates occur. Viewpoint checks, local route repair, and trajectory validation avoid queries on accumulated global point clouds. Long-range planning scales with retained task descriptors and the topological graph. When no valid local viewpoint remains, historical retrieval selects at most $K$ tasks for ordering, yielding an $O(K^2)$ cost matrix. Execution uses a locally validated initial route segment.

The map representation and planning results in Section~\ref{sec:system_efficiency} support the practical benefit of this allocation: compact history retains the tasks and return connectivity needed to sustain exploration beyond the local map. Scalability follows from preserving the information required for future decisions at an appropriate granularity, with current geometry supplying the detail required for flight.

\section{Conclusion}
\label{sec:conclusion}

This paper presented \method{}, a hierarchical spatio-temporal map representation for mission-scale UAV exploration without persistent, globally queryable dense metric geometry. The robot-centric local map supports geometric reasoning. At the \textit{observation level}, LiDAR-keyframe visibility records preserve prior sensing evidence and suppress repeated frontiers. At the \textit{task level}, the cascaded frontier representation carries unfinished objectives across map shifts, with odometry anchors and topological routes supporting their recovery. Local-first, anchor-associated planning retrieves distant tasks and reconstructs executable viewpoints and routes from current observations. Task context therefore determines where exploration should resume, while current geometry determines how recovered tasks are executed. This design fixes dense-map capacity to the local-window size and confines exploration-dependent memory growth to compact observation context and task context. Across three large-scale simulated scenes, \method{} reduces the memory footprint of the map representation by 94.00--99.00\% relative to EPIC while achieving the shortest completion time in every scene, reducing exploration time by 31.98--41.88\%. The compact mission state also supports responsive topology maintenance and global planning throughout these missions. In onboard forest and complex construction-site deployments, \method{} sustains autonomous exploration over a combined 507.40~m trajectory while reducing the memory footprint of the map representation by 92.67\% and 62.42\%, respectively, relative to an explicit point-cloud ikd-tree. Together, these results establish that compact mission history can preserve long-range exploration continuity while dense geometric state remains bounded by the local map. By combining observation context for frontier reasoning with task context for recovering unfinished objectives, \method{} provides a scalable map representation for large-scale autonomous exploration under constrained onboard resources.

\printcredits

\section{Declaration of interests}
The authors declare that they have no known competing financial interests or personal relationships that could have appeared to influence the work reported in this paper.

\section{Acknowledgment}
This study was jointly supported by the National Natural Science Foundation Project (No. 42201477, No. 42130105) and Wuhan Natural Science Foundation Project (No. 2025041001010363).

\renewcommand{\bibfont}{\fontsize{8pt}{9.5pt}\selectfont}
\bibliographystyle{cas-model2-names}
\bibliography{reference}

\end{sloppypar}
\end{document}